\documentclass[letterpaper]{article} 
\usepackage{aaai2026}  
\usepackage{times}  
\usepackage{helvet}  
\usepackage{courier}  
\usepackage[hyphens]{url}  
\usepackage{graphicx} 
\usepackage{natbib}  
\usepackage{caption} 
\usepackage{algorithm}
\usepackage{algorithmic}

\usepackage{newfloat}
\usepackage{listings}
\DeclareCaptionStyle{ruled}{labelfont=normalfont,labelsep=colon,strut=off} 
\floatstyle{ruled}
\newfloat{listing}{tb}{lst}{}
\floatname{listing}{Listing}
\usepackage{aaai2026}
\usepackage{times}
\usepackage{helvet}
\usepackage[hyphens]{url}  
\usepackage{courier}
\usepackage{natbib}  

\usepackage{multirow}
\usepackage[utf8]{inputenc} 
\usepackage[T1]{fontenc}    
\usepackage{url}            
\usepackage{booktabs}       
\usepackage{amsfonts}       
\usepackage{nicefrac}       
\usepackage{microtype}      
\usepackage{xcolor}         
\usepackage{times}
\usepackage{epsfig}
\usepackage{graphicx}
\usepackage{amsmath}
\usepackage{amssymb}
\usepackage{indentfirst} 
\usepackage{changepage} 
\usepackage{soul}
\usepackage{amsthm}
\usepackage{booktabs}
\usepackage{algorithm}
\usepackage{algorithmic}
\usepackage{makecell}
\usepackage{appendix}
\usepackage{subfigure}
\usepackage{fancyhdr}
\usepackage{siunitx}
\usepackage{makecell}
\usepackage{amssymb}
\usepackage{pifont}
\usepackage{xcolor} 

\makeatletter
\@ifundefined{isChecklistMainFile}{
  \newif\ifreproStandalone
  \reproStandalonetrue
}{
  \newif\ifreproStandalone
  \reproStandalonefalse
}
\makeatother

\usepackage{times}
\usepackage{helvet}
\usepackage{courier}
\usepackage{xcolor}
\usepackage{pifont}
\usepackage{fontawesome} 
\usepackage{bbding}
\usepackage{blindtext}
\newcommand*\samethanks[1][\value{footnote}]{\footnotemark[#1]}

\usepackage{bibentry}
 
\begin{document}

\setlength{\leftmargini}{20pt}
\makeatletter\def\@listi{\leftmargin\leftmargini \topsep .5em \parsep .5em \itemsep .5em}
\def\@listii{\leftmargin\leftmarginii \labelwidth\leftmarginii \advance\labelwidth-\labelsep \topsep .4em \parsep .4em \itemsep .4em}
\def\@listiii{\leftmargin\leftmarginiii \labelwidth\leftmarginiii \advance\labelwidth-\labelsep \topsep .4em \parsep .4em \itemsep .4em}\makeatother

\setcounter{secnumdepth}{0}
\renewcommand\thesubsection{\arabic{subsection}}
\renewcommand\labelenumi{\thesubsection.\arabic{enumi}}

\newcounter{checksubsection}
\newcounter{checkitem}[checksubsection]

\newcommand{\checksubsection}[1]{%
  \refstepcounter{checksubsection}%
  \paragraph{\arabic{checksubsection}. #1}%
  \setcounter{checkitem}{0}%
}

\newcommand{\checkitem}{%
  \refstepcounter{checkitem}%
  \item[\arabic{checksubsection}.\arabic{checkitem}.]%
}
\newcommand{\question}[2]{\normalcolor\checkitem #1 #2 \color{blue}}
\newcommand{\ifyespoints}[1]{\makebox[0pt][l]{\hspace{-15pt}\normalcolor #1}}

%
\title{Spikingformer: A Key Foundation  Model for Spiking Neural Networks}

\author{Chenlin Zhou\textsuperscript{\rm 1, \rm 2}, Liutao Yu$^{2}$, Zhaokun Zhou$^{1}$,  Han Zhang$^{2,3}$, Jiaqi Wang$^{2,3}$,  \\  Huihui Zhou$^{2}$, Zhengyu Ma$^{2}$\thanks{Corresponding author},  Yonghong Tian$^{1,2,4}$\samethanks\\
}
\affiliations {
    \textsuperscript{\rm 1}School of Electronic and Computer Engineering, Shenzhen Graduate School, Peking University\\
    \textsuperscript{\rm 2}Pengcheng Laboratory \textsuperscript{\rm 3}Harbin Institute of Technology
    \textsuperscript{\rm 4}School of Computer Science, Peking University\\
    chenlinzhou25@stu.pku.edu.cn, mazhy@pcl.ac.cn, yhtian@pku.edu.cn
}

\maketitle
\begin{abstract}
\begin{quote}
Spiking neural networks (SNNs) offer a promising energy-efficient alternative to artificial neural networks, due to their event-driven spiking computation. 
However, some foundation SNN backbones (including Spikformer and SEW ResNet) suffer from non-spike computations (integer-float multiplications) caused by the structure of their residual connections. These non-spike computations increase SNNs' power consumption and make them unsuitable for deployment on mainstream neuromorphic hardware. 
In this paper, we analyze the spike-driven behavior of the residual connection methods in SNNs. We then present Spikingformer, a novel spiking transformer backbone that merges the MS Residual connection with Self-Attention in a biologically plausible way to address the non-spike computation challenge in Spikformer while maintaining global modeling capabilities.
We evaluate Spikingformer across 13 datasets spanning large static images, neuromorphic data, and natural language tasks, and demonstrate the effectiveness and universality of Spikingformer, setting a vital benchmark for spiking neural networks.
In addition, with the spike-driven features and global modeling capabilities, Spikingformer is expected to become a more efficient general-purpose SNN backbone towards energy-efficient artificial intelligence.
Code: \url{https://github.com/TheBrainLab/Spikingformer}

\end{quote}
\end{abstract}

\section{Introduction}

\begin{table*}[htb]
\begin{center}
\begin{tabular}{lccccc}
\toprule
{Settings} & SEW-ResNet & MS-ResNet & Spikformer & SD-Transformer & Spikingformer$^\dagger$ \\
\midrule
Test Resolution & 224$\times$224 & 224$\times$224 / 288$\times$288 & 224$\times$224  &  288$\times$288 & 224$\times$224 \\ 
Model Size & 77.28M & 78.37M & 66.34M &  66.34M & 66.34M \\ 
Model Type &  CNN  & CNN & Transformer & Transformer  & Transformer  \\ 
Attention &  No  & No & SSA & SDSA  & PSSA  \\ 
Global Attention &  \ding{55}  & \ding{55} & \ding{51} & \ding{55}  & \ding{51}  \\ 
Residual Connection &  SEW Residual   & MS Residual  & SEW Residual   & MS Residual   & MS Residual   \\ 
Spike-Driven &  \ding{55}  & \ding{51} & \ding{55} & \ding{51}  & \ding{51}  \\ 
\midrule
ImageNet-1K Acc ($\%$) &  69.26 & 74.21 / 76.02 & 74.81 & 77.07 &  \textbf{77.64} \\   
\bottomrule
\end{tabular}
\end{center}
\vspace{-0.3cm}
\caption{Comparison with other foundation backbones in the SNN domain, including SEW-ResNet \cite{fang2021deep}, MS-ResNet \cite{hu2021advancing}, Spikformer \cite{zhou2023spikformer}, SD-Transformer \cite{yao2023spike}.}
\vspace{-0.2cm}
\label{complexity}
\end{table*}

Being regarded as the third generation of neural networks \cite{maass1997networks}, the brain-inspired Spiking Neural Networks (SNNs) are potential competitors to Artificial Neural Networks (ANNs) due to their high biological plausibility, event-driven property, and low power consumption on neuromorphic hardware \cite{roy2019towards}. In particular, the utilization of binary spike signals allows SNNs to adopt low-power accumulation (AC) instead of the traditional high-power multiply-accumulation (MAC), leading to significant energy efficiency gains and making SNNs increasingly popular \cite{chen2023training}.

As SNNs go deeper, their performance has improved significantly \cite{hu2018residual,fang2021deep, zheng2021going, hu2021advancing}.
ResNet with a skipping connection has been extensively studied to extend the depth of SNNs \cite{fang2021deep,zhou2023spikformer}. 
SEW ResNet \cite{fang2021deep}, a representative Convolutional Neural Network (CNN) based SNN, easily implements identity mapping by Spike-Element-Wise (SEW) Residual connection and overcomes the problems of vanishing/exploding gradients of Spiking ResNet \cite{hu2021spiking}. SEW ResNet is the first deep SNN directly trained with more than 100 layers.
MS-ResNet \cite{hu2021advancing} proposed the Membrane-based Shortcut (MS) Residual connection for spiking neural networks, effectively addressing gradient explosion and vanishing problems without performance degradation in CNN-based SNNs. 
Spikformer \cite{zhou2023spikformer}, a directly trained representative transformer-based SNN with SEW Residual connection, is proposed by leveraging both self-attention capability and biological properties of SNNs. It is the first successful exploration of applying flourishing transformer architecture into SNN design, and shows powerful performance. 
However, Spikformer faces the challenge of non-spike computations (integer-float multiplications) caused by the SEW Residual connection. This not only limits their ability to fully leverage the benefits of event-driven processing in energy efficiency, but also makes it difficult to deploy and optimize their performance on neuromorphic hardware \cite{chen2023training}.

We carry out an in-depth comparison between the SEW Residual connection and the MS Residual connection. 
Then, we creatively combine MS Residual and Self-Attention in a spike-driven way to address the challenge of non-spike computation in Spikformer while achieving superior performance.
%
We name this model Spikingformer, in contrast to Spike-Driven Transformer (SD-Transformer) \cite{yao2023spike}, which combines MS Residual with linear attention yet lacks global modeling capability, limiting its potential as a general-purpose SNN backbone.
The main foundation SNN backbones are shown in Tab. \ref{complexity}. Spikingformer is the backbone that contains both spike-driven features and global modeling capabilities. Our contributions are as follows:

1) We conduct a systematic comparison between MS Residual connection and SEW Residual connection, analyzing their spike-driven behavior, firing rate, and performance.

2) We develop a novel spiking transformer model, named Spikingformer, which innovatively integrates MS Residual and Self-Attention in a spike-driven way. Furthermore, leveraging both spike-driven computation and global attention modeling capabilities, Spikingformer is expected to serve as a foundation model for general-purpose energy-efficient artificial intelligence.

3) We validate Spikingformer on 13 datasets across static image classification, neuromorphic classification, and natural language understanding, establishing a vital experimental benchmark for the SNN community.

\section{Related Work}
The non-differentiability of spiking neurons in SNNs makes it difficult to train SNNs directly, but a common solution is to employ the surrogate gradient for backpropagation \cite{neftci2019surrogate}. Existing direct training SNNs can be roughly divided into two categories: convolution-based SNNs and transformer-based SNNs.
\subsection{Convolution-based Spiking Neural Network}
%
In the field of direct training, SNNs are unfolded over simulation time steps and trained with backpropagation through time \cite{lee2016training,shrestha2018slayer}. Due to the non-differentiability of spiking neurons, surrogate gradient method is employed for backpropagation \cite{lee2020enabling,neftci2019surrogate}.  SEW ResNet\cite{fang2021deep} is a representative convolution-based SNN model by direct training, and is the first to increase the number of layers in SNNs to be larger than 100. However, the ADD gate in residual connections of SEW ResNet produces non-spike computations of integer-float multiplications in deep convolution layers. \cite{chen2023training} has identified the problem of non-spike computations in SEW ResNet and Spikformer, and attempts to solve it through adding an auxiliary accumulation pathway during training and removing it during inference. This strategy needs tedious extra operations and results in a significant performance degradation compared with the original models. 

\subsection{Transformer-based Spiking Neural Network.} 
Most existing SNNs borrow architectures from convolutional neural networks (CNNs), so their performance is limited by the performance of CNNs. The transformer architecture, originally designed for natural language processing \cite{vaswani2017attention}, has achieved great success in many computer vision tasks, including image classification \cite{dosovitskiy2020image,yuan2021tokens}, object detection \cite{carion2020end,zhu2020deformable}, and semantic segmentation \cite{wang2021pyramid,yuan2021volo}. The structure of the transformer leads to a novel kind of SNN, with great potential to break through the bottleneck of SNNs' performance. So far, two main related works: Spikformer \cite{zhou2023spikformer} and Spikeformer \cite{li2022spikeformer}, have proposed spiking neural networks based on transformer architecture. Although Spikeformer replaces the activation function used in the feedforward layers with a spiking activation function, there are still a lot of non-spike operations remaining, including floating point multiplication, division, and exponential operation. Spikformer proposes a novel Spiking Self Attention (SSA) module by using spike-form Query, Key, and Value without softmax, and achieves state-of-the-art performances on many datasets. 
However, the structure of Spikformer with residual connection still contains non-spike computation.
%
In our study, we innovatively integrates MS Residual and Self-Attention in a spike-driven way to address this problem.


\section{Methods}
\subsection{Spiking Neuron Model}
Spiking neuron is the fundamental unit of SNNs, we choose  Leaky Integrate-and-Fire (LIF) model as the spike neuron in our work. The dynamics can be formulated as follows:
\begin{gather}
H[t]=V[t-1]+\frac{1}{\tau}\left(X[t]-\left(V[t-1]-V_{\text {reset }}\right)\right),\\
S[t]=\Theta\left(H[t]-V_{t h}\right) , \\
V[t]=H[t](1-S[t])+V_{\text {reset }} S[t] ,
\end{gather}
where $\tau$ is the membrane time constant, and $X[t]$ is the input current at time step $t$. When the membrane potential $H[t]$ exceeds the firing threshold $V_{th}$, the spiking neuron will trigger a spike $S[t]$. $\Theta(v)$ is the Heaviside step function, which equals to 1 when $v\geq 0$ and 0 otherwise. $V[t]$ represents the membrane potential after the triggered event, which equals to $H[t]$ if no spike is generated and otherwise equals to the reset potential $V_{reset}$.  

\subsection{Spike-Driven Behavior in SNN Residual Learning}
There are mainly three types of residual connections in SNNs: Vanilla
Residual \cite{zheng2021going}, SEW Residual \cite{fang2021deep}, MS residual \cite{hu2021advancing}. However, Vanilla Residual connection suffers from performance degradation and gradient vanishing/exploding. Therefore, we mainly discuss the latter two residual connections.

\begin{figure}[tbp]
	\centering
	\includegraphics[width=0.45\textwidth]{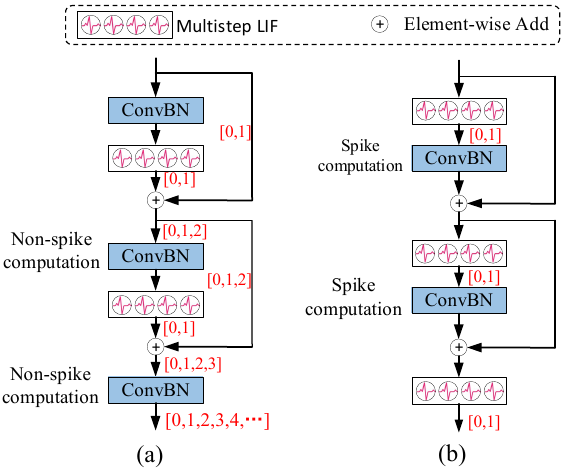}
    \vspace{-0.4cm}
    \caption{The residual learning in Spikformer and Spikingformer.
    (a) shows the SEW Residual learning of Spikformer, which contains non-spike computation (integer-float multiplications) in $\operatorname{ConvBN}$ layer.
    (b) shows the MS Residual connection, which is adopted in Spikingformer. 
    MS Residual could effectively avoid integer-float multiplications, following the spike-driven principle. 
   }
	\label{fig: Neuromorphic residual}
	\vspace{-0.3cm}
\end{figure}

\textbf{The SEW Residual adopted in Spikformer.}
At present,  Spikformer \cite{zhou2023spikformer} is the representative work combining deep SNNs with transformer architecture.
The residual learning plays an extremely important role in Spikformer, but the SEW Residual connections in Spikformer and SEW ResNet lead to non-spike computation (integer-float multiplications), which are not event-driven computations. 
As shown in Fig.\ref{fig: Neuromorphic residual}(a),  the residual learning of Spikformer and SEW ResNet could be formulated as follows:
\begin{equation}
\left\{
\begin{aligned}
&\textbf{O}_l=\operatorname{SN}_l(\operatorname{ConvBN}_l(\textbf{O}_{l-1})) + \textbf{O}_{l-1}, \\
&\textbf{O}_{l+1}=\operatorname{SN}_{l+1}(\operatorname{ConvBN}_{l+1}(\textbf{O}_{l})) + \textbf{O}_{l},
\end{aligned}
\right.
\end{equation}
we denotes $\operatorname{SN}_l(\operatorname{ConvBN}_l(\textbf{O}_{l-1}))$ as $\textbf{S}_l$. This residual design inevitably brings in non-spike data and thus MAC operations in the next layer/block. In particular, $\textbf{S}_l$ and $\textbf{O}_{l-1}$ are spike signals, and their output $\textbf{O}_l$ are non-spike signal whose range is $\{0, 1, 2\}$. Non-spike data destructs event-driven computation in the next convolution layer when computing $\textbf{S}_{l+1}$ of $\textbf{O}_{l+1}$. As the depth of the network increases, the range of non-spike data values transmitted to the deeper layer of the network will also expand. In our implementations of Spikformer, the range of the non-spike data could increase to $\{0,1,2,...,16\}$ when testing Spikormer-8-512 on ImageNet 2012. Obviously, the range of non-spike data is approximately proportional to the number of residual blocks in Spikformer and SEW ResNet. 
In fact, integer-float multiplications are usually implemented in the same way as floating-point multiplication in hardware. In this case, the network will incur high energy consumption, approaching to the energy consumption of ANNs with the same structure, which is unacceptable for SNNs.

\textbf{The MS Residual adopted in Spikingformer.}
Fig.\ref{fig: Neuromorphic residual}(b) shows the MS Residual connection adopted in Spikingformer. It could effectively avoid floating-point multiplications and integer-float multiplications, following the spike-driven principle. The spike-driven residual learning could be easily formulated as follows:
\begin{equation}
\left\{
\begin{aligned}
&\textbf{O}_l=\operatorname{ConvBN}_l(\operatorname{SN}_l(\textbf{O}_{l-1})) + \textbf{O}_{l-1}, \\
&\textbf{O}_{l+1}=\operatorname{ConvBN}_{l+1}(\operatorname{SN}_{l+1}(\textbf{O}_{l})) + \textbf{O}_{l},
\end{aligned}
\right.
\end{equation}
%
In this structure, we denote $\operatorname{ConvBN}_l(\operatorname{SN}_l(\textbf{O}_{l-1}))$ as $\textbf{S}_l$. $\textbf{S}_l + \textbf{O}_{l-1}$ belongs to the floating point addition operation, which is the same as the addition operation in $\operatorname{SN}$ layer. The floating point addition operation is the most essential operation of SNN. 
Obviously, the output $\textbf{O}_{l}$ is also floating point and will pass through an $\operatorname{SN}$ layer before participating in the next $\operatorname{ConvBN}$ computation. 
Therefore, the pure spike-form feature will be generated after the processing of $\operatorname{SN}$ layer, and the computation of  $\operatorname{ConvBN}$ layer will become the floating point addition operation, following the spike-driven principle and reducing energy consumption vastly. 
In addition, we show the impact of SEW Residual and MS Residual on the firing behavior and performance of transformer-based SNNs, respectively, in the experimental section.

\subsection{Spikingformer}
\begin{figure*}[htb]
	\centering
	\includegraphics[width=0.85\textwidth]{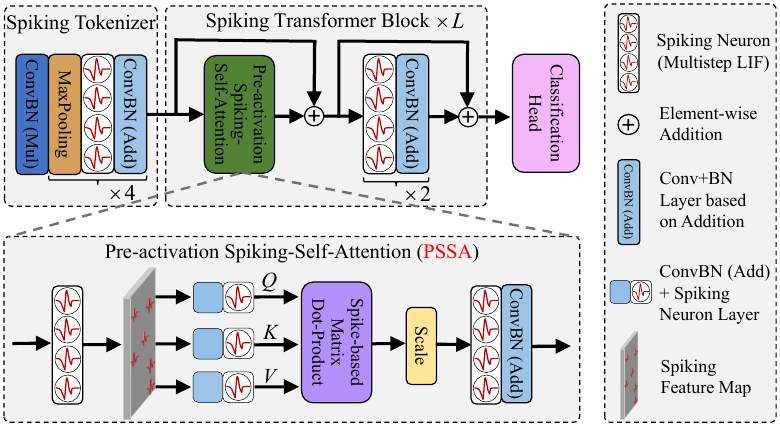}
	\vspace{-0.1cm}
	\caption{The overview of Spikingformer, which consists of a Spiking Tokenizer, several Spiking Transformer Blocks, and a Classification Head. Note that Mutistep LIF is the Leaky Integrate-and-Fire (LIF) neuron model \cite{fang2021deep, zhou2023spikformer} with time steps $T>1$. Same with Spikformer,  $T$ is an independent dimension for the spike neuron layer. In other layers, it is merged with the batch size. We use $\operatorname{ConvBN}$ to represent a convolution layer and its subsequent BN layer in this work.}
	\label{fig:spikingformer}
	\vspace{-0.2cm}
\end{figure*}
Spikingformer family contains two foundation models: Spikingformer and Spikingformer$^\dagger$. Spikingformer is a novel and pure transformer-based spiking neural network through integrating spike-driven MS Residual blocks. Spikingformer$^\dagger$ is a variant of Spikingformer. The pipeline of Spikingformer is shown in Fig.\ref{fig:spikingformer}. 

\textbf{Overall Architecture}. 
Our proposed Spikingformer contains a  Spiking Tokenizer (ST), several Spiking Transformer Blocks, and a Classification Head. Given a 2D image sequence $\textbf{I} \in \mathbb{R}^{T \times C \times H \times W}$ (Note that $C \text{=} 3$ in static datasets like ImageNet 2012, $C \text{=} 2$ in neuromorphic datasets like DVS-Gesture), we use the Spiking Tokenizer block for downsampling and patch embedding, where the inputs can be projected as spike-form patches $\textbf{X} \in \mathbb{R}^{T \times N \times D}$. Obviously, the first layer of Spiking Tokenizer also play a spike encoder role when taking static images as input. After Spiking Tokenizer, 
the spiking patches $\textbf{X}_{0}$ will pass to the $L$ Spiking Transformer Blocks.
Similar to the standard ViT encoder block, a Spiking Transformer Block contains a Spiking Self Attention (SSA) \cite{zhou2023spikformer} and a Spiking MLP block. In the last, a fully-connected-layer (FC) is used for the Classification Head. Note that we use a global average-pooling (GAP) before the fully-connected layer to reduce the parameters of FC and improve the classification capability of Spikingformer.
\begin{align}
&\textbf{X} =\operatorname{ST}(\textbf{I}), \quad \textbf{I} \in \mathbb{R}^{T \times C \times H \times W}, \textbf{X} \in \mathbb{R}^{T \times N \times D} \\
&\textbf{X}_l^{\prime}=\operatorname{SSA}\left(\textbf{X}_{l-1}\right)+\textbf{X}_{l-1},  \textbf{X}_l^{\prime} \in \mathbb{R}^{T \times N \times D},  \\
&\textbf{X}_l=\operatorname{SMLP}\left(\textbf{X}_l^{\prime}\right)+\textbf{X}_l^{\prime},  \textbf{X}_l \in \mathbb{R}^{T \times N \times D}, \\
&\textbf{Y}=\operatorname{FC}\left(\operatorname{GAP}\left(\textbf{X}_L\right)\right).  
\end{align}

\textbf{Spiking Tokenizer.} 
As shown in Fig.\ref{fig:spikingformer}, Spiking Tokenizer mainly contains two functions: 1) convolutional spiking patch embedding, and 2) downsampling to project the feature map into a smaller fixed size. The spiking patch embedding is similar to the convolutional stream in Vision Transformer \cite{xiao2021early,hassani2021escaping}, where the dimension of spike-form feature channels gradually increases in each convolution layer and finally matches the embedding dimension of patches. In addition, the first layer of Spiking Tokenizer is utilized as a spike encoder when using static images as input. As shown in Eq. \ref{eq:spe1} and  Eq. \ref{eq:spe2}, the convolution part of $\operatorname{ConvBN}$ represents the 2D convolution layer (stride-1, 3 × 3 kernel size). $\operatorname{MP}$ and $\operatorname{SN}$ represent maxpooling (stride-2) and mutistep spiking neuron, respectively. Eq. \ref{eq:spe1} is used for Spiking Patch Embedding without Downsampling (SPE), Eq. \ref{eq:spe2} is Spiking Patch Embedding with Downsampling (SPED). We could use multiple SPEs or SPEDs for specific classification tasks with different downsampling requirements. For example, we use 4 SPEDs for ImageNet 2012 dataset classification with input size as 224*224 (using 16 times downsampling). we use 2 SPEs and 2 SPEDs for CIFAR dataset classification with input size as 32*32 (using 4 times downsampling). After the processing of the Spiking Tokenizer block, the input $\textbf{I}$ is split into an image patch sequence $\textbf{X} \in \mathbb{R}^{T \times N \times D}$.
\begin{align}
&\textbf{I}_{i}= \operatorname{ConvBN}(\operatorname{SN}(\textbf{I})), \label{eq:spe1}\\
&\textbf{I}_{i}= \operatorname{ConvBN}(\operatorname{MP}(\operatorname{SN}(\textbf{I}))). \label{eq:spe2}
\end{align}

\textbf{Spiking Transformer Block.} A Spiking Transformer Block contains a Pre-activation Spiking Self-Attention (PSSA) block and a Spiking MLP block. PSSA retains Spiking Self-Attention (SSA)'s global modeling capabilities while making some modifications to be spike-driven and more generalized. Thus, PSSA can be seen as an important variant of SSA. The modifications include: 1) We change the spiking neuron layer position according to our proposed spike-driven residual mechanism, avoiding the multiplication of integers and floating-point weights. 2) To enhance generalization across diverse tasks, we choose $\operatorname{ConvBN}$ in place of $\operatorname{LinearBN}$ (linear layer and batch normalization) in Spikformer. Therefore, the PSSA can be formulated as follows:
\begin{gather}
\textbf{X}^{\prime} = \operatorname{SN}(\textbf{X}),
\end{gather}
\begin{equation}
\left\{
\begin{aligned}
\textbf{Q} &= \operatorname{SN}_{Q}(\operatorname{ConvBN}_{Q}(\textbf{X}^{\prime})), \\
\textbf{K} &= \operatorname{SN}_{K}(\operatorname{ConvBN}_{K}(\textbf{X}^{\prime})), \\
\textbf{V} &= \operatorname{SN}_{V}(\operatorname{ConvBN}_{V}(\textbf{X}^{\prime})),
\end{aligned}
\right.
\end{equation}
\begin{gather}
\operatorname{Attention}(\textbf{Q}, \textbf{K}, \textbf{V})= \operatorname{ConvBN}(\operatorname{SN}(\textbf{Q} \textbf{K}^{\mathrm{T}} \textbf{V} * s)),
\end{gather}
where $\textbf{Q}, \textbf{K}, \textbf{V} \in \mathbb{R}^{T \times N \times D}$ are pure spike data (only containing 0 and 1). $s$ is the scaling factor as in \cite{zhou2023spikformer}, controlling the large value of the matrix multiplication
result. The Spiking MLP block consists of two SPEs, which are formulated in Eq.\ref{eq:spe1}. Spiking Transformer Block is shown in Fig.\ref{fig:spikingformer}, and it is the main component of Spikingformer.

\begin{table*}[!htb]
  \centering
  \vspace{+0.2cm}
    \begin{tabular}{llccp{1.3cm}<{\centering}p{1.4cm}<{\centering}p{1.5cm}<{\centering}p{1.5cm}<{\centering}}
    \toprule
    \multicolumn{1}{l}{Methods} & \multicolumn{1}{l}{Architecture} & Param{} (M)&Train Size &Test Size & {Time Step} & Energy (mJ)  & {Top-1 Acc ($\%$)} \\
    \midrule
    Spiking ResNet &ResNet-34 &21.79 &224$^2$ &224$^2$ & 350 & {59.30}  & 71.61 \\
    Spiking ResNet &ResNet-50 &25.56 &224$^2$ &224$^2$ & 350 & {70.93}  & 72.75 \\
    SEW ResNet & SEW ResNet-34 &21.79 &224$^2$ &224$^2$ &4 &{4.04}  & 67.04 \\
    SEW ResNet & SEW ResNet-50 &25.56 &224$^2$ &224$^2$ &4 &{4.89}  & 67.78 \\
    SEW ResNet & SEW ResNet-101 &44.55 &224$^2$ &224$^2$ &4 &{8.91}  & 68.76 \\
    SEW ResNet & SEW ResNet-152 &60.19 &224$^2$  &224$^2$ &4 &{12.89}  & 69.26 \\
    MS-ResNet & ResNet-104 &78.37 &224$^2$ &224$^2$ &5 &-  &74.21\\
    MS-ResNet & ResNet-104 &78.37 &224$^2$ &288$^2$ &5 &-  &76.02\\
Spikformer   &{Spikformer-8-384} &16.81 &224$^2$ &224$^2$ &4 &{12.43}  & 70.24\\
Spikformer  &{Spikformer-8-512}&29.68 &224$^2$ &224$^2$ & 4 &{18.82}  & {73.38}\\
Spikformer  &{Spikformer-8-768}&66.34 &224$^2$ &224$^2$ & 4 &{32.07}  & {74.81}\\
    {SD-Transformer}  & {S-Transformer-8-384} &16.81 &224$^2$ &224$^2$  &4 &3.90  & {72.28}\\
    {SD-Transformer}  &{S-Transformer-8-512}  &29.68 &224$^2$ &224$^2$  & 4 &4.50  & {74.57}\\
    {SD-Transformer}  &{S-Transformer-8-768} &66.34 &224$^2$ &288$^2$  & 4 &6.09  & {77.07}\\
    {ANN}   & {Transformer-8-512} & {29.68} &224$^2$ &224$^2$ & {1} & {38.34}  & {{80.80}} \\
    \midrule
    {\multirow{3}{*}{\textbf{Spikingformer}}}
       &{Spikingformer-8-384} &16.81 &224$^2$ &224$^2$ &4 &{4.69}  & {72.45}\\
    &{Spikingformer-8-512}&29.68 &224$^2$ &224$^2$ & 4 &{7.46}  & {74.79}\\
    &{Spikingformer-8-768}&66.34 &224$^2$ &224$^2$ & 4 &{13.68}  & {75.85}\\
    \cmidrule(lr){2-8} 
    {\multirow{3}{*}{\textbf{Spikingformer$^\dagger$}}}
    &{Spikingformer-8-384} &16.81 &224$^2$ &224$^2$ &4 &{5.61}  & {74.35}\\
    &{Spikingformer-8-512}&29.68 &224$^2$ &224$^2$ & 4 &{8.68}  & {76.54}\\
    &{Spikingformer-8-768}&66.34 &224$^2$ &224$^2$ & 4 &{16.30}  & \textbf{77.64}\\
    \bottomrule
    \end{tabular}%
  \caption{Results on ImageNet-1k classification. Power is calculated as the average theoretical energy consumption of an image inference on ImageNet, whose detail is shown in Eq.\ref{eq:energy1}. Same as Spikformer, our Spikingformer-$L$-$D$ represents a Spikingformer model with $L$ spiking transformer blocks and $D$ feature embedding dimensions. }
  \vspace{-4mm}
  \label{tab:imagenet}%
\end{table*}%

\textbf{Classification Head.} We use a fully-connected-layer as the classifier behind the last Spiking Transformer Block. In detail, the classifier could be realized in four forms:  $\operatorname{AvgPooling}\text{-}\operatorname{FC}$, $\operatorname{SN}\text{-}\operatorname{AvgPooling}\text{-}\operatorname{FC}$, $\operatorname{FC}\text{-}\operatorname{AvgPooling}$, $\operatorname{SN}\text{-}\operatorname{FC}\text{-}\operatorname{AvgPooling}$ :
\begin{align}
&\textbf{Y}=\operatorname{FC}(\operatorname{AvgPooling}(\textbf{X}_L)) \label{eq:class1}, \\
&\textbf{Y}=\operatorname{FC}(\operatorname{AvgPooling}(\operatorname{SN}(\textbf{X}_L))) \label{eq:class2} ,\\
&\textbf{Y}=\operatorname{AvgPooling}(\operatorname{FC}(\textbf{X}_L)) \label{eq:class3}, \\
&\textbf{Y}=\operatorname{AvgPooling}(\operatorname{FC}(\operatorname{SN}(\textbf{X}_L))). \label{eq:class4}
\end{align}
$\operatorname{AvgPooling}$ after $\operatorname{FC}$ (like $\operatorname{SN}\text{-}\operatorname{FC}\text{-}\operatorname{AvgPooling}$, $\operatorname{FC}\text{-}\operatorname{AvgPooling}$) could be considered as computing the average of neuron firing, a post-processing of network, but in this way $\operatorname{FC}$ usually requires numerous parameters. $\operatorname{AvgPooling}$ before $\operatorname{FC}$ (like $\operatorname{AvgPooling}\text{-}\operatorname{FC}$, $\operatorname{SN}\text{-}\operatorname{AvgPooling}\text{-}\operatorname{FC}$) could effectively reduce parameters compared with the previous ways. Only $\operatorname{SN}\text{-}\operatorname{FC}\text{-}\operatorname{AvgPooling}$ could avoid floating-point multiplication operation, but it needs more $\operatorname{FC}$ parameters than $\operatorname{AvgPooling}\text{-}\operatorname{FC}$ or $\operatorname{SN}\text{-}\operatorname{AvgPooling}\text{-}\operatorname{FC}$. In addition, it also hinders the classification ability of the network. In this work, we mainly adopt the way of $\operatorname{AvgPooling}$ ahead of $\operatorname{FC}$, and choose $\operatorname{AvgPooling}\text{-}\operatorname{FC}$ as the classifier of Spikingformer by default. Some experimental analysis on the classification head will be discussed in Appendix.

\textbf{Spikingformer$^\dagger$}. As a key variant, Spikingformer$^\dagger$ is the Spikingformer with ConvBN-MaxPool-LIF (CML) downsampling \cite{zhou2023enhancing}. CML can overcome the imprecision problem of gradient backpropagation in Spikformer or Spikingformer, which improves network performance while reducing computational cost at the same time.

\subsection{Theoretical Energy Consumption Calculation}\label{sec：energy}
The homogeneity of convolution allows the following BN and linear scaling transformation to be fused into the convolutional layer with an added bias when deployment \cite{ding2019acnet, ding2021repvgg, hu2021spiking, chen2023training}. Therefore, when calculating the theoretical energy consumption, the consumption of BN layers could be ignored. 
We calculate the number of synaptic operations of spikes before calculating the theoretical energy consumption.
\begin{equation}
SOP^l=f r \times T \times {FLOPs}^l,
\end{equation}
where $l$ is a block/layer in Spikingformer, $fr$ is the firing rate of the block/layer, and $T$ is the simulation time step of the spike neuron. ${FLOPs}^l$ refers to floating point operations of block/layer $l$, which is the number of multiply-and-accumulate (MAC) operations. And ${SOP}^l$ is the number of spike-based accumulate (AC) operations.
We estimate the theoretical energy consumption of Spikingformer according to \cite{kundu2021hire,hu2021advancing,horowitz20141,kundu2021spike,panda2020toward}. We assume that the MAC and AC operations are implemented on the 45nm hardware \cite{horowitz20141}, where $E_{MAC}=4.6pJ$ and $E_{AC}=0.9pJ$. The theoretical energy consumption of Spikingformer can be calculated as follows:
\begin{multline}
E_{Snn}=E_{A C} \times\left(\sum_{i=2}^N S O P_{\operatorname{Conv} }^i+\sum_{j=1}^M S O P_{\operatorname{SSA}}^j\right)+ \\
E_{M A C} \times\left(FLOP_{\operatorname{Conv}}^1\right), \label{eq:energy1}
\end{multline}
Eq.\ref{eq:energy1} shows the energy consumption of Spikingformer. $FLOP_{Conv}^1$ is the first layer encoding the non-spike input into spike-form. Then the SOPs of $N$ SNN Conv layers and $M$ SSA layers are added together and multiplied by $E_{AC}$. 



\begin{table*}[!htb]
  \centering
  \vspace{+0.2cm}
    \begin{tabular}{lp{1.1cm}<{\centering}p{0.1cm}<{\centering}p{1.1cm}<{\centering}p{1.1cm}<{\centering}p{0.1cm}<{\centering}p{1.1cm}<{\centering}p{1.1cm}<{\centering}p{0.1cm}<{\centering}p{1.1cm}<{\centering}p{1.1cm}<{\centering}p{0.1cm}<{\centering}p{1.2cm}<{\centering}}
    \toprule
    \multicolumn{1}{c}{\multirow{2}[4]{*}{Method}} & \multicolumn{3}{c}{CIFAR10} & \multicolumn{3}{c}{CIFAR100} & \multicolumn{3}{c}{DVS128} & \multicolumn{3}{c}{CIFAR10-DVS} \\
\cmidrule(l{2pt}r{2pt}){2-4}\cmidrule(l{2pt}r{2pt}){5-7}\cmidrule(l{2pt}r{2pt}){8-10}\cmidrule(l{2pt}r{2pt}){11-13}
    &{Param} & {$T$} & {Acc} &Param & {$T$} & {Acc} &Param & {$T$} & {Acc} &Param & {$T$} & {Acc}\\
    \midrule
SEW ResNet  & $-$ & $-$ & $-$ & $-$ & $-$ & $-$ & $-$ & 16 & {74.4} & $-$ & {16} & {97.9} \\
MS-ResNet  &$-$ & $-$ & 91.72 & $-$  & $-$ & $66.83$  & $-$ & $-$ & 75.6 & $-$ & $-$ & $-$  \\
Spikformer   &5.76 & 4 & 94.80 & 5.76  & 4    & 76.95  & 2.57  & 10  & 95.8  & 2.57  & 10  & 78.6 \\
Spikformer   &9.32 & 4 & 95.51 & 9.32  & 4    & 78.21  & 2.57  & 16  & 97.9  & 2.57  & 16  & 80.9 \\
SD-Transformer  &   10.28    & 4  &  95.60  &   10.28    &   4    &  78.4      &    2.57   &   16    &  {99.3}     &   2.57    &  16     & 80.0 \\
STSA  &  $-$  &$-$ & $-$   &  $-$    &  $-$   & $-$     &    1.99   &   16    &  98.7     &   1.99    &  16     & 79.93 \\
{QKFormer}  &   6.74    & 4  &  {96.18}  &   6.74    &   4    &  {81.15}      &    1.50   &   16    &  98.6     &   1.50    &  16     & {84.0} \\

\cmidrule(lr){2-13}
Transformer (ANN)  &  9.32  &1 & 96.73   &  9.32    &  1   & 81.02   &    $-$   &   $-$    &  $-$     &   $-$    &  $-$     & $-$ \\

\midrule
\multirow{2}{*}{\textbf{Spikingformer}}   &  5.76    & 4  &  95.22  &   5.76    &   4    &  78.34      &    2.57   &   10    &  96.2     &   2.57    &  10     & 79.9 \\
    &   9.32    & 4  &  95.81  &   9.32    &   4    &  79.21      &    2.57   &   16    &  98.3     &   2.57    &  16     & 81.3 \\
\cmidrule(lr){2-13}
\multirow{2}{*}{\textbf{Spikingformer$^\dagger$}}   &   5.76    & 4  &  95.54  &   5.76    &   4    &  78.87   &    2.57   &   10   &  97.2    & 2.57  &  10  & 80.5 \\
    &   9.32    & 4  &  \textbf{95.95}  &   9.32    &   4    &  \textbf{80.37}   &    2.57   &   16   &  \textbf{98.6}    & 2.57  &  16  & \textbf{81.4} \\
    \bottomrule
    \end{tabular}%
    \caption{Comparision on CIFAR10, CIFAR100, DVS128 and CIFAR10-DVS. "Param" denotes "Parameter (M)", "Acc" denotes "Top-1 Accuracy (\%)", and "$T$" denotes "Time Step".} 
    \vspace{-2mm}
  \label{tab:small_dataset}%
\end{table*}%

\section{Experiments}
In this section, we carry out experiments on the static dataset ImageNet \cite{deng2009imagenet}, the static dataset CIFAR \cite{krizhevsky2009learning} (including CIFAR10 and  CIFAR100), the neuromorphic datasets (including CIFAR10-DVS and DVS128-Gesture \cite{amir2017dvsg}), and the natural language understanding tasks (GLUE datasets) to evaluate Spikingformer. 
The models for conducting experiments are implemented based on SpikingJelly \cite{SpikingJelly}.

\subsection{ImageNet-1k Classification}\label{sec:Ic}
We compared Spikingformer with Spiking ResNet \cite{hu2018residual}, SEW ResNet \cite{fang2021deep}, MS-ResNet\cite{hu2021advancing}, Spikformer \cite{zhou2023spikformer}, SD-Transformer  \cite{yao2023spike} on ImageNet-1k, which is shown in Tab. \ref{tab:imagenet}.
%
Note that we recalculate the energy consumption of Spikformer in Appendix because the non-spike computation of Spikformer can not be directly calculated by Eq.\ref{eq:energy1}.
Spikingformer-8-512 achieves 74.79$\%$ top-1 classification accuracy on ImageNet using 4 time steps, significantly outperforms Spikformer-8-512 by 1.41$\%$, outperforms MS-ResNet model by 0.58$\%$ and outperforms SEW ResNet-152 model by 5.53$\%$. Spikingformer-8-512 is with 7.463 mJ theoretical energy consumption, which reduces energy consumption by 60.36$\%$, compared with 18.819 mJ of Spikformer-8-512.
Spikingformer-8-768 achieves 75.85$\%$ top-1 classification accuracy on ImageNet using 4 time steps, significantly outperforms Spikformer-8-768 by 1.04$\%$, outperforms the MS-ResNet model by 1.64$\%$ and outperforms SEW ResNet-152 model by 6.59$\%$. Spikingformer-8-768 is with 13.678 mJ theoretical energy consumption, which reduces energy consumption by 57.34$\%$, compared with 32.074 mJ of Spikformer-8-768. 
%
%
%
Spikingformer$^\dagger$-8-512 achieves 76.54$\%$ accuracy, which outperforms Spikformer-8-512 by 3.16$\%$ and outperforms SD-Transformer-8-512 by 1.97$\%$. 
Spikingformer$^\dagger$-8-768 achieves 77.64$\%$ Top-1 classification accuracy, which outperforms Spikformer-8-768 by 2.83$\%$. Compared with other foundation SNN backbones (SEW-ResNet, MS-ResNet, Spikformer, SD-Transformer), Spikingformer achieves the best performance in ImageNet-1k due to its spike-driven features and global modeling capabilities.


\begin{table*}[htb]
\begin{center}
\begin{small}
\centering
\setlength{\tabcolsep}{1.7mm}
\fontsize{9.3pt}{\baselineskip}\selectfont
\begin{tabular}{lp{1.5cm}<{\centering}p{0.8cm}<{\centering}cccccccc|p{1.8cm}<{\centering}}
\midrule
{Model} & {Energy}$_\text{ (mJ)}$ & {Time}  & {MNLI}$_\text{-m/mm}$ & {QQP} & {QNLI} & {SST-2} & {CoLA} & {STS-B} & {MRPC}$_\text{F1}$ & {RTE} & {Avg. Acc ($\%$)} \\ 
\midrule
BERT$_\texttt{base}$  & 51.41 & --   & 84.6/83.4 & 71.2 & 90.5 & 93.5 & 52.1 & 85.8 & 88.9 & 66.4 & 79.6 \\
Q2BERT                & --    & --   & 47.2/47.3 & 67.0 & 61.3 & 80.6 & 0.0  & 4.7  & 81.2 & 52.7 & 49.1 \\
ELMo                   & --    & --   & 68.6/--   & 86.2 & 71.1 & 91.5 & 44.1 & 70.4 & 76.6 & 53.4 & 70.2 \\
SpikeBERT             & 14.30 & 4    & 71.4/71.0 & 68.2 & 66.4 & 85.4 & 16.9 & 18.7 & 82.0 & 57.5 & 59.7 \\
\midrule
\textbf{Spikingformer}         & 6.76    & 4    & 71.9/72.5 & 84.7 & 76.0 & 87.2 & 24.4 & 54.5 & 79.7 & 55.6 & \textbf{66.8} \\
\toprule
\end{tabular}
\end{small}
\end{center}
\vskip -0.1in
\caption{The results on the  Natural Language Understanding task (GLUE datasets). "Avg. Acc" denotes "Average Accuracy".}
\label{glue}
\vskip -0.1in
\end{table*}

\subsection{CIFAR and Neuromorphic Tasks}
The results are shown in Tab. \ref{tab:small_dataset}. We compared Spikingformer with  SEW ResNet \cite{fang2021deep},  MS-ResNet \cite{hu2021advancing}, Spikformer \cite{zhou2023spikformer}, SD-Transformer \cite{yao2023spike}, STSA \cite{ijcai2023p344}, QKFormer \cite{zhou2024qkformer}. 


\textbf{CIFAR Classification.}
From the results, 
We find that the performance of Spikingformer models surpass all the models of Spikformer with the same number of parameters. 
In CIFAR10, our Spikingformer achieves 95.81$\%$ classification accuracy, significantly outperforms Spikformer by 0.30$\%$ and outperforms MS-ResNet-482 by 3.91$\%$. Spikingformer$^\dagger$ achieves 95.95$\%$ accuracy and outperforms SD-Transformer by 0.35$\%$.
In CIFAR100, Spikingformer achieves 79.21$\%$ classification accuracy, significantly outperforms Spikformer by 1.00$\%$ and outperforms MS-ResNet by 12.38$\%$. Spikingformer$^\dagger$ achieves 80.37$\%$ accuracy and outperforms SD-Transformer by 1.97$\%$.
%
%
Transformer (ANN) is only 0.69$\%$ and 1.00$\%$ higher than Spikingformer$^\dagger$ on CIFAR10 and CIFAR100. 

\textbf{Neuromorphic Classification.}
 We compare our method with SOTA methods on DVS-Gesture. In detail, we adopt four SPEDs in the Spiking Tokenizer block due to the 128*128 image size of CIFAR10-DVS and adopt 2 spiking transformer blocks with 256 patch embedding dimension. The number of time steps of the spiking neuron is 10 or 16. The number of training epochs is 106, which is the same as Spikformer. The learning rate is initialized to 0.1 and decayed with a cosine schedule.
The results of CIFAR10-DVS are shown in Tab.\ref{tab:small_dataset}. Spikingformer achieves 81.3$\%$ top-1 accuracy with 16 time steps and 79.9$\%$ accuracy with 10 time steps, significantly outperforms Spikformer by 0.4$\%$ and 1.3$\%$ respectively. 
Spikingformer$^\dagger$ achieves 81.4$\%$ accuracy with 16 time steps and outperforms SD-Transformer by 1.4$\%$. 
%
We compare our method with SOTA methods on CIFAR10-DVS in Tab.\ref{tab:small_dataset}. Spikingformer achieves 98.3$\%$ top-1 accuracy with 16 time steps and 96.2$\%$ accuracy with 10 time steps, outperforms Spikformer by 0.4$\%$ and 0.4$\%$ respectively.
%

\subsection{Natural Language Understanding}\label{sec:NLU}
We evaluate Spikingformer on the standard GLUE (General Language Understanding Evaluation) benchmark \cite{wang2018glue}, which is a widely adopted collection of datasets designed to evaluate and advance natural language understanding (NLU) capabilities in machine learning models. GLUE contains 8 subsets for classification and regression, including single-sentence classification (CoLA, SST-2), pairwise sentence comparison (MPRC, QQP, RTE), and natural language inference (STS-B, MNLI, QNLI, WNLI). 
We pretrain Spikingformer on Wikipedia-English \cite{devlin2019bert}  by masked language pretraining \cite{devlin2019bert}, then finetune on GLUE dev set. The experimental results are shown in Tab. \ref{glue}.

We compare Spikingformer with BERT$_\texttt{base}$ \cite{devlin2019bert}, Q2BERT \cite{zhang2020ternarybert}, ELMo \cite{sarzynska2021detecting}, SpikeBERT \cite{lv2023spikebert}.
Our Spikingformer achieves 66.8$\%$ average accuracy, which outperforms SpikeBERT by 7.1 $\%$. More analysis see Appendix.


\begin{figure}[!ht]
    \centering
    \subfigure[The spike behavior of Spikformer-8-512 on ImageNet]{\includegraphics[width=1.0\linewidth]{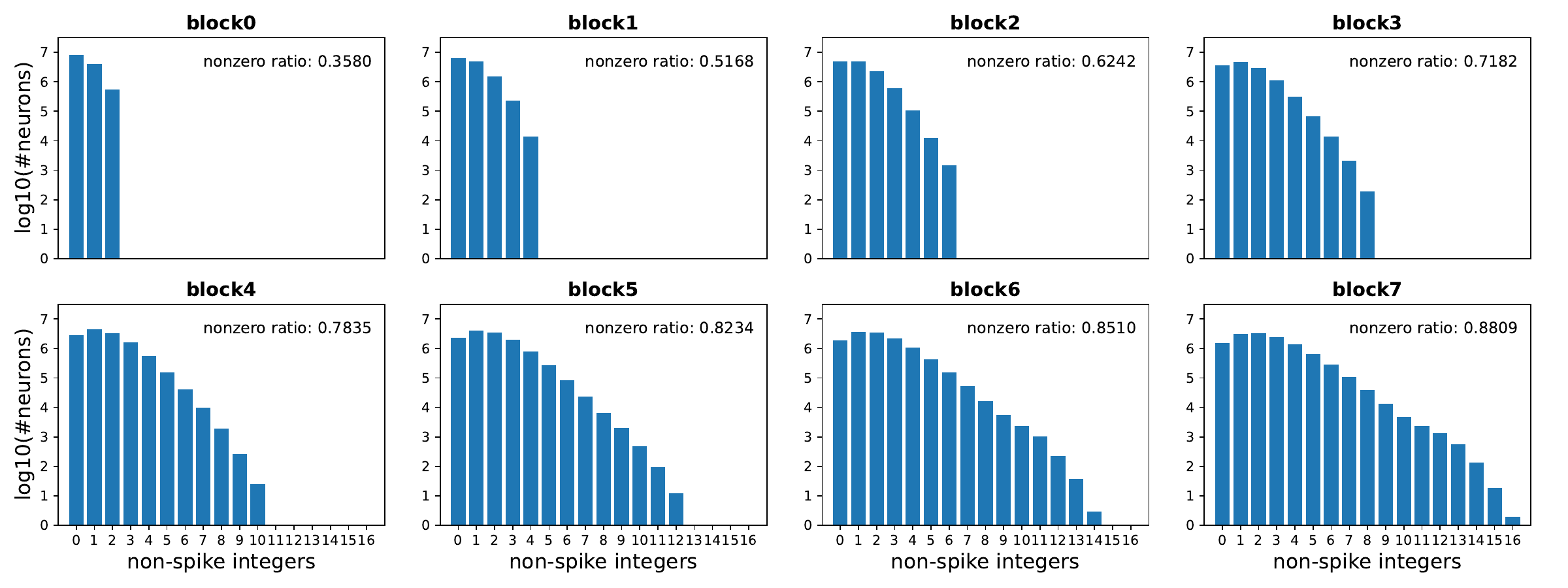} \label{fig:nonspike}}
    \vspace{-4mm}
    \\
    \subfigure[The spike behavior of Spikingformer-8-512 on ImageNet]{\includegraphics[width=1.0\linewidth]{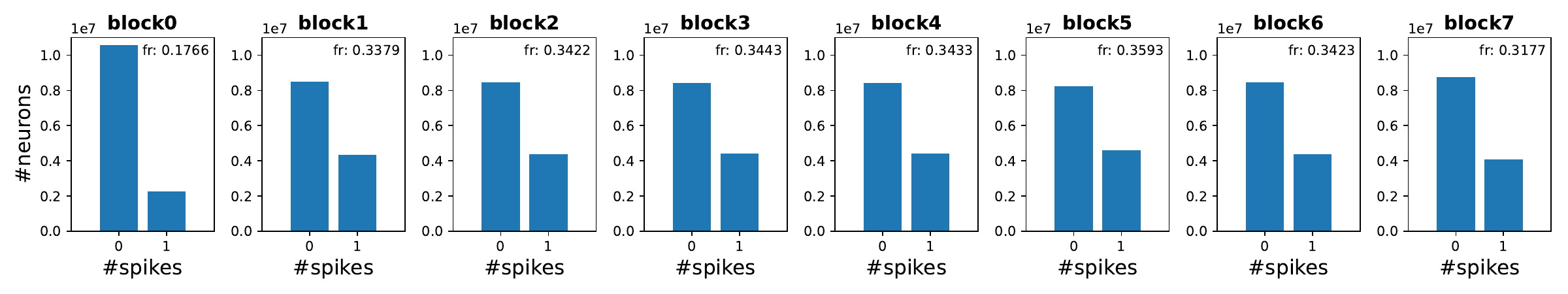} \label{fig:spike}}
    \vspace{-4mm}
    \caption{The spike-driven behavior of Spikingformer and Spiformer.
    \subref{fig:nonspike} Histogram of the input data of each block in Spikformer-8-512. The abscissa means non-spike data range with $\{0, 1, 2,..., 16\}$ before $\operatorname{Conv}$ layer in the transformer block of Spikformer. 
    The nonzero ratio indicates the ratio of non-zero input units for each block.
    \subref{fig:spike} Histogram of the input data of each block in Spikingformer-8-512. The abscissa means binary spike data with $\{0, 1\}$ before $\operatorname{Conv}$ layer in the transformer block of Spikingformer. The ordinate means of the neuron numbers of $\{0, 1\}$.
   }
    \label{fig: spike and nonspike}
    \vspace{-4mm}
\end{figure}

\subsection{Discussion}
In this part, we conduct a systematic comparison between MS Residual connection and SEW Residual connection on Transformer-based SNN (corresponding to Spikingformer and Spikformer, respectively), analyzing their spike-driven behavior, power consumption, and performance metrics from an experimental perspective.

\textbf{Spike Behavior Visualization}. We visualize the spike behavior of spikingformer and spikformer in Fig. \ref{fig: spike and nonspike}. Spikformer-8-512 has 8 transformer blocks and two residual connections in every transformer block. Thus, the non-spike numbers can be accumulated up to 16 in Spikformer-8-512. 
In addition, the results show Spikingformer could effectively avoid integer-float multiplications common in Spikformer. In addition, $fr$ in Fig.\ref{fig:spike} represents the firing rate of the input data for each spiking transformer block of Spikingformer. We observe that Spikingformer have a lower firing rate on ImageNet compared with Spikformer (Fig.\ref{fig:nonspike}), which further reduces synaptic operations and thus energy consumption.

\textbf{Energy Consumption and Performance Impact}. The results can be seen in Tab \ref{tab:imagenet}. Compared to Spikformer, Spikingformer achieves higher performance while significantly reducing energy consumption. The primary reason for this energy reduction is Spikingformer's ability to effectively eliminate integer-float multiplications. Additionally, a lower firing rate on ImageNet further contributes to the improved energy efficiency.

\section{Conclusion}
In this work, we propose Spikingformer, a fully spike-driven transformer-based spiking neural network that innovatively integrates MS Residual connections with Self-Attention in a biologically plausible, spike-driven manner. This design effectively overcomes the non-spike computation limitations of existing spiking transformers, such as Spikformer, while maintaining global modeling capabilities. 
Extensive experiments across 13 diverse datasets, including large-scale image classification, neuromorphic data classification, and natural language understanding tasks, demonstrate that Spikingformer consistently outperforms previous foundation SNN backbones. With its efficient spike-driven computation and global modeling capacity, Spikingformer establishes itself as a robust, general-purpose SNN backbone and a key benchmark for the SNN community.

\section*{Acknowledgements}
The study was funded by the National Natural Science Foundation of China under contracts No. 62425101, No. 62332002, No. 62027804, No.62088102, and the major key project of the Pengcheng Laboratory (PCL2025A02).

\bibliography{aaai2026}

@article{zhou2023enhancing,
  title={Enhancing the performance of transformer-based spiking neural networks by SNN-optimized downsampling with precise gradient backpropagation},
  author={Zhou, Chenlin and Zhang, Han and Zhou, Zhaokun and Yu, Liutao and Ma, Zhengyu and Zhou, Huihui and Fan, Xiaopeng and Tian, Yonghong},
  journal={arXiv preprint arXiv:2305.05954},
  year={2023}
}

@inproceedings{zhu2015aligning,
  title={Aligning books and movies: Towards story-like visual explanations by watching movies and reading books},
  author={Zhu, Yukun and Kiros, Ryan and Zemel, Rich and Salakhutdinov, Ruslan and Urtasun, Raquel and Torralba, Antonio and Fidler, Sanja},
  booktitle={Proceedings of the IEEE international conference on computer vision},
  pages={19--27},
  year={2015}
}

@article{xing2024spikelm,
  title={Spikelm: Towards general spike-driven language modeling via elastic bi-spiking mechanisms},
  author={Xing, Xingrun and Zhang, Zheng and Ni, Ziyi and Xiao, Shitao and Ju, Yiming and Fan, Siqi and Wang, Yequan and Zhang, Jiajun and Li, Guoqi},
  journal={arXiv preprint arXiv:2406.03287},
  year={2024}
}

@article{neftci2019surrogate,
  title={Surrogate gradient learning in spiking neural networks: Bringing the power of gradient-based optimization to spiking neural networks},
  author={Neftci, Emre O and Mostafa, Hesham and Zenke, Friedemann},
  journal={IEEE Signal Processing Magazine},
  volume={36},
  number={6},
  pages={51--63},
  year={2019},
  publisher={IEEE}
}

@article{zhang2020ternarybert,
  title={Ternarybert: Distillation-aware ultra-low bit bert},
  author={Zhang, Wei and Hou, Lu and Yin, Yichun and Shang, Lifeng and Chen, Xiao and Jiang, Xin and Liu, Qun},
  journal={arXiv preprint arXiv:2009.12812},
  year={2020}
}

@article{sarzynska2021detecting,
  title={Detecting formal thought disorder by deep contextualized word representations},
  author={Sarzynska-Wawer, Justyna and Wawer, Aleksander and Pawlak, Aleksandra and Szymanowska, Julia and Stefaniak, Izabela and Jarkiewicz, Michal and Okruszek, Lukasz},
  journal={Psychiatry Research},
  volume={304},
  pages={114135},
  year={2021},
  publisher={Elsevier}
}

@inproceedings{devlin2019bert,
  title={Bert: Pre-training of deep bidirectional transformers for language understanding},
  author={Devlin, Jacob and Chang, Ming-Wei and Lee, Kenton and Toutanova, Kristina},
  booktitle={Proceedings of the 2019 conference of the North American chapter of the association for computational linguistics: human language technologies, volume 1 (long and short papers)},
  pages={4171--4186},
  year={2019}
}

@article{wang2018glue,
  title={GLUE: A multi-task benchmark and analysis platform for natural language understanding},
  author={Wang, Alex and Singh, Amanpreet and Michael, Julian and Hill, Felix and Levy, Omer and Bowman, Samuel R},
  journal={arXiv preprint arXiv:1804.07461},
  year={2018}
}

@article{lv2023spikebert,
  title={SpikeBERT: A Language Spikformer Learned from BERT with Knowledge Distillation},
  author={Lv, Changze and Li, Tianlong and Xu, Jianhan and Gu, Chenxi and Ling, Zixuan and Zhang, Cenyuan and Zheng, Xiaoqing and Huang, Xuanjing},
  journal={arXiv preprint arXiv:2308.15122},
  year={2023}
}

@Article{zhu2020deformable,
  author  = {Zhu, Xizhou and Su, Weijie and Lu, Lewei and Li, Bin and Wang, Xiaogang and Dai, Jifeng},
  title   = {Deformable DETR: Deformable Transformers for End-to-End Object Detection},
  journal = {arXiv preprint arXiv:2010.04159},
  year    = {2020},
}

@InProceedings{wang2021pyramid,
  author    = {Wang, Wenhai and Xie, Enze and Li, Xiang and Fan, Deng-Ping and Song, Kaitao and Liang, Ding and Lu, Tong and Luo, Ping and Shao, Ling},
  booktitle = {Proceedings of the IEEE/CVF International Conference on Computer Vision (ICCV)},
  title     = {Pyramid vision transformer: A versatile backbone for dense prediction without convolutions},
  pages     = {568--578},
  year      = {2021},
}

@Article{hu2021advancing,
  author  = {Hu, Yifan and Wu, Yujie and Deng, Lei and Li, Guoqi},
  title   = {Advancing residual learning towards powerful deep spiking neural networks},
  journal = {arXiv preprint arXiv:2112.08954},
  year    = {2021},
}

@Article{krizhevsky2009learning,
  author    = {Krizhevsky, Alex},
  title     = {Learning multiple layers of features from tiny images},
  journal   = {},
  publisher = {Citeseer},
  year      = {2009},
}

@InProceedings{ding2019acnet,
  author    = {Ding, Xiaohan and Guo, Yuchen and Ding, Guiguang and Han, Jungong},
  booktitle = {Proceedings of the IEEE/CVF international conference on computer vision},
  title     = {Acnet: Strengthening the kernel skeletons for powerful cnn via asymmetric convolution blocks},
  pages     = {1911--1920},
  year      = {2019},
}

@InProceedings{shrestha2018slayer,
  author    = {Shrestha, Sumit B and Orchard, Garrick},
  booktitle = {Proceedings of the International Conference on Neural Information Processing Systems (NeurIPS)},
  title     = {Slayer: Spike layer error reassignment in time},
  volume    = {31},
  year      = {2018},
}

@InProceedings{yuan2021tokens,
  author    = {Yuan, Li and Chen, Yunpeng and Wang, Tao and Yu, Weihao and Shi, Yujun and Jiang, Zi-Hang and Tay, Francis EH and Feng, Jiashi and Yan, Shuicheng},
  booktitle = {Proceedings of the IEEE/CVF International Conference on Computer Vision (ICCV)},
  title     = {Tokens-to-token vit: Training vision transformers from scratch on imagenet},
  pages     = {558--567},
  year      = {2021},
}

@InProceedings{vaswani2017attention,
  author    = {Vaswani, Ashish and Shazeer, Noam and Parmar, Niki and Uszkoreit, Jakob and Jones, Llion and Gomez, Aidan N and Kaiser, {\L}ukasz and Polosukhin, Illia},
  booktitle = {Proceedings of the International Conference on Neural Information Processing Systems (NeurIPS)},
  title     = {Attention is all you need},
  volume    = {30},
  year      = {2017},
}

@InProceedings{xiao2021early,
  author    = {Xiao, Tete and Singh, Mannat and Mintun, Eric and Darrell, Trevor and Doll{\'a}r, Piotr and Girshick, Ross},
  booktitle = {Proceedings of the International Conference on Neural Information Processing Systems (NeurIPS)},
  title     = {Early convolutions help transformers see better},
  pages     = {30392--30400},
  volume    = {34},
  year      = {2021},
}

@Article{lee2016training,
  author    = {Lee, Jun Haeng and Delbruck, Tobi and Pfeiffer, Michael},
  title     = {Training deep spiking neural networks using backpropagation},
  pages     = {508},
  volume    = {10},
  journal   = {Frontiers in neuroscience},
  publisher = {Frontiers Media SA},
  year      = {2016},
}

@Article{hu2021spiking,
  author    = {Hu, Yangfan and Tang, Huajin and Pan, Gang},
  title     = {Spiking deep residual networks},
  journal   = {IEEE Transactions on Neural Networks and Learning Systems},
  publisher = {IEEE},
  year      = {2021},
}

@Article{panda2020toward,
  author    = {Panda, Priyadarshini and Aketi, Sai Aparna and Roy, Kaushik},
  title     = {Toward scalable, efficient, and accurate deep spiking neural networks with backward residual connections, stochastic softmax, and hybridization},
  pages     = {653},
  volume    = {14},
  journal   = {Frontiers in Neuroscience},
  publisher = {Frontiers Media SA},
  year      = {2020},
}

@InProceedings{carion2020end,
  author       = {Carion, Nicolas and Massa, Francisco and Synnaeve, Gabriel and Usunier, Nicolas and Kirillov, Alexander and Zagoruyko, Sergey},
  booktitle    = {Proceedings of the European Conference on Computer Vision (ECCV)},
  title        = {End-to-end object detection with transformers},
  organization = {Springer},
  pages        = {213--229},
  year         = {2020},
}

@Article{maass1997networks,
  author    = {Maass, Wolfgang},
  title     = {Networks of spiking neurons: the third generation of neural network models},
  number    = {9},
  pages     = {1659--1671},
  volume    = {10},
  journal   = {Neural networks},
  publisher = {Elsevier},
  year      = {1997},
}

@InProceedings{deng2009imagenet,
  author    = {Deng, Jia and Dong, Wei and Socher, Richard and Li, Li-Jia and Li, Kai and Fei-Fei, Li},
  booktitle = {Proceedings of the IEEE/CVF Conference on Computer Vision and Pattern Recognition (CVPR)},
  title     = {Imagenet: A large-scale hierarchical image database},
  pages     = {248--255},
  year      = {2009},
}

@InProceedings{kundu2021hire,
  author    = {Kundu, Souvik and Pedram, Massoud and Beerel, Peter A},
  booktitle = {Proceedings of the IEEE/CVF International Conference on Computer Vision (ICCV)},
  title     = {Hire-snn: Harnessing the inherent robustness of energy-efficient deep spiking neural networks by training with crafted input noise},
  pages     = {5209--5218},
  year      = {2021},
}

@Article{chen2023training,
  author  = {Chen, Guangyao and Peng, Peixi and Li, Guoqi and Tian, Yonghong},
  title   = {Training Full Spike Neural Networks via Auxiliary Accumulation Pathway},
  journal = {arXiv preprint arXiv:2301.11929},
  year    = {2023},
}

@Article{hu2018residual,
  author  = {Hu, Yangfan and Tang, Huajin and Pan, Gang},
  title   = {Spiking Deep Residual Networks},
  doi     = {10.1109/TNNLS.2021.3119238},
  pages   = {1-6},
  journal = {IEEE Transactions on Neural Networks and Learning Systems},
  year    = {2021},
}

@InProceedings{zhou2023spikformer,
  author    = {Zhaokun Zhou and Yuesheng Zhu and Chao He and Yaowei Wang and Shuicheng YAN and Yonghong Tian and Li Yuan},
  booktitle = {The Eleventh International Conference on Learning Representations},
  title     = {Spikformer: When Spiking Neural Network Meets Transformer},
  url       = {https://openreview.net/forum?id=frE4fUwz_h},
  year      = {2023},
}

@Article{yuan2021volo,
  author  = {Yuan, Li and Hou, Qibin and Jiang, Zihang and Feng, Jiashi and Yan, Shuicheng},
  title   = {Volo: Vision outlooker for visual recognition},
  journal = {arXiv preprint arXiv:2106.13112},
  year    = {2021},
}

@Article{hassani2021escaping,
  author  = {Hassani, Ali and Walton, Steven and Shah, Nikhil and Abuduweili, Abulikemu and Li, Jiachen and Shi, Humphrey},
  title   = {Escaping the big data paradigm with compact transformers},
  journal = {arXiv preprint arXiv:2104.05704},
  year    = {2021},
}

@InProceedings{amir2017dvsg,
  author    = {Amir, Arnon and Taba, Brian and Berg, David and Melano, Timothy and McKinstry, Jeffrey and Di Nolfo, Carmelo and Nayak, Tapan and Andreopoulos, Alexander and Garreau, Guillaume and Mendoza, Marcela and Kusnitz, Jeff and Debole, Michael and Esser, Steve and Delbruck, Tobi and Flickner, Myron and Modha, Dharmendra},
  booktitle = {Proceedings of the IEEE/CVF Conference on Computer Vision and Pattern Recognition (CVPR)},
  title     = {A Low Power, Fully Event-Based Gesture Recognition System},
  pages     = {7243--7252},
  year      = {2017},
}

@InProceedings{horowitz20141,
  author       = {Horowitz, Mark},
  booktitle    = {2014 IEEE International Solid-State Circuits Conference Digest of Technical Papers (ISSCC)},
  title        = {1.1 computing's energy problem (and what we can do about it)},
  organization = {IEEE},
  pages        = {10--14},
  year         = {2014},
}

@Article{li2022spikeformer,
  author  = {Li, Yudong and Lei, Yunlin and Yang, Xu},
  title   = {Spikeformer: A Novel Architecture for Training High-Performance Low-Latency Spiking Neural Network},
  journal = {arXiv preprint arXiv:2211.10686},
  year    = {2022},
}

@InProceedings{ding2021repvgg,
  author    = {Ding, Xiaohan and Zhang, Xiangyu and Ma, Ningning and Han, Jungong and Ding, Guiguang and Sun, Jian},
  booktitle = {Proceedings of the IEEE/CVF conference on computer vision and pattern recognition},
  title     = {Repvgg: Making vgg-style convnets great again},
  pages     = {13733--13742},
  year      = {2021},
}

@Article{lee2020enabling,
  author    = {Lee, Chankyu and Sarwar, Syed Shakib and Panda, Priyadarshini and Srinivasan, Gopalakrishnan and Roy, Kaushik},
  title     = {Enabling spike-based backpropagation for training deep neural network architectures},
  pages     = {119},
  volume    = {14},
  journal   = {Frontiers in neuroscience},
  publisher = {Frontiers},
  year      = {2020},
}

@InProceedings{dosovitskiy2020image,
  author    = {Dosovitskiy, Alexey and Beyer, Lucas and Kolesnikov, Alexander and Weissenborn, Dirk and Zhai, Xiaohua and Unterthiner, Thomas and Dehghani, Mostafa and Minderer, Matthias and Heigold, Georg and Gelly, Sylvain and others},
  booktitle = {International Conference on Learning Representa- tions (ICLR)},
  title     = {An image is worth 16x16 words: Transformers for image recognition at scale},
  year      = {2020},
}

@InProceedings{kundu2021spike,
  author    = {Kundu, Souvik and Datta, Gourav and Pedram, Massoud and Beerel, Peter A},
  booktitle = {Proceedings of the IEEE/CVF Winter Conference on Applications of Computer Vision (WACV)},
  title     = {Spike-thrift: Towards energy-efficient deep spiking neural networks by limiting spiking activity via attention-guided compression},
  pages     = {3953--3962},
  year      = {2021},
}

@InProceedings{zheng2021going,
  author    = {Hanle Zheng and Yujie Wu and Lei Deng and Yifan Hu and Guoqi Li},
  booktitle = {Proceedings of the AAAI Conference on Artificial Intelligence (AAAI)},
  title     = {{Going Deeper With Directly-Trained Larger Spiking Neural Networks}},
  pages     = {11062--11070},
  year      = {2021},
}

@Article{roy2019towards,
  author  = {Roy, Kaushik and Jaiswal, Akhilesh and Panda, Priyadarshini},
  title   = {Towards Spike-based Machine Intelligence With Neuromorphic Computing},
  number  = {7784},
  pages   = {607--617},
  volume  = {575},
  journal = {Nature},
  year    = {2019},
}

@InProceedings{fang2021deep,
  author    = {Fang, Wei and Yu, Zhaofei and Chen, Yanqi and Huang, Tiejun and Masquelier, Timoth and Tian, Yonghong},
  booktitle = {Proceedings of the International Conference on Neural Information Processing Systems (NeurIPS)},
  title     = {{Deep Residual Learning in Spiking Neural Networks}},
  pages     = {21056--21069},
  volume    = {34},
  year      = {2022},
}

@article{zhou2024qkformer,
  title={Qkformer: Hierarchical spiking transformer using qk attention},
  author={Zhou, Chenlin and Zhang, Han and Zhou, Zhaokun and Yu, Liutao and Huang, Liwei and Fan, Xiaopeng and Yuan, Li and Ma, Zhengyu and Zhou, Huihui and Tian, Yonghong},
  journal={arXiv preprint arXiv:2403.16552},
  year={2024}
}

@article{yao2023spike,
  title={Spike-driven transformer},
  author={Yao, Man and Hu, Jiakui and Zhou, Zhaokun and Yuan, Li and Tian, Yonghong and Xu, Bo and Li, Guoqi},
  journal={Advances in neural information processing systems},
  volume={36},
  pages={64043--64058},
  year={2023}
}

@inproceedings{ijcai2023p344,
  title     = {Spatial-Temporal Self-Attention for Asynchronous Spiking Neural Networks},
  author    = {Wang, Yuchen and Shi, Kexin and Lu, Chengzhuo and Liu, Yuguo and Zhang, Malu and Qu, Hong},
  booktitle = {Proceedings of the Thirty-Second International Joint Conference on
               Artificial Intelligence, {IJCAI-23}},
  publisher = {International Joint Conferences on Artificial Intelligence Organization},
  editor    = {Edith Elkind},
  pages     = {3085--3093},
  year      = {2023},
  month     = {8},
  note      = {Main Track},
  doi       = {10.24963/ijcai.2023/344},
  url       = {https://doi.org/10.24963/ijcai.2023/344},
}

@misc{SpikingJelly,
	title = {SpikingJelly},
	author = {Fang, Wei and Chen, Yanqi and Ding, Jianhao and Chen, Ding and Yu, Zhaofei and Zhou, Huihui and Timothée Masquelier and Tian, Yonghong and other contributors},
	year = {2020},
	howpublished = {\url{https://github.com/fangwei123456/spikingjelly}},
	note = {Accessed: YYYY-MM-DD},
}

\newpage
\section{Appendix}



\subsection{Theoretical Analysis in Fusion of Convolution and Batch Normalization}\label{sec:convbn}
Obviously, the binary spikes become floating-point values when it has passed through the convolution layer, which leads to subsequent MAC operations in Batch Normalization (BN) layer. 
However, the homogeneity of convolution allows the following BN and linear scaling transformation to be equivalently fused into the convolutional layer with an added bias when deployment. In particular, each BN layer and its preceding convolution layer are fused into a convolution $\operatorname{ConvBN}$ layer with a bias vector. The kernel and bias $\{ W, B\}$ of $\operatorname{ConvBN}$ can be calculated from $\{W, \mu, \sigma, \gamma, \beta \}$. 
The process of convolution and batch normalization for an input element $x_{i}$ can be formulated as follow:
\begin{gather}
y_{\operatorname{Conv}}=w_{\operatorname{Conv}} \cdot x_i+b_{\operatorname{Conv}} 
\end{gather}

\begin{multline}
y_i=\operatorname{B N}_{\gamma, \beta}\left(x_i\right)=\gamma \frac{x_i-\mu_{\text { }}}{\sqrt{\sigma_{\text { }}^2+\varepsilon}}+\beta \\
=\frac{\gamma}{\sqrt{\sigma_{\text { }}^2+\varepsilon}} x_i+\beta-\frac{\gamma \cdot \mu_{\text { }}}{\sqrt{\sigma_{\text { }}^2+\varepsilon}}
\end{multline}
Thus in deployment, the batch normalization could be formulated as: $y_{\operatorname{BN}}=w_{\operatorname{BN}} \cdot x_{i}+b_{\operatorname{\operatorname{BN}}}$. Therefore, the above steps could be fused:
\begin{multline}
y_i = w_{\operatorname{BN}}(w_{\operatorname{Conv}} \cdot x_{i} +b_{\operatorname{Conv}}) + b_{\operatorname{BN}} \\ 
= w_{\operatorname{BN}} \cdot w_{\operatorname{Conv}} \cdot x_{i} + w_{\operatorname{BN}}b_{\operatorname{Conv}}+b_{\operatorname{BN}}
\end{multline}
The equivalent convolution layer $\operatorname{ConBN}$: $W=w_{\operatorname{BN}} \cdot w_{\operatorname{Conv}}$; $B=w_{\operatorname{BN}} \cdot b_{\operatorname{Conv}} + b_{\operatorname{BN}}$. 

\subsection{Multi-head Spiking Self-Attention in Spikingformer}
The Multi-head Spiking Self-Attention (MSSA) can be easily  formulated  as follows:

\begin{equation}
X^{\prime}=\operatorname{SN}(X) 
\end{equation}
\begin{multline}
Q, K, V=\operatorname{S N}_q\left({\operatorname{ConvBN}}_q\left(X^{\prime}\right)\right), \\ 
\operatorname{S N}_k\left({\operatorname{ConvBN}_{k}}\left(X^{\prime}\right)\right), \operatorname{S N}_v\left({\operatorname{ConvBN}}_v\left(X^{\prime}\right)\right)
\end{multline}
\begin{multline}
Q^{\prime}, K^{\prime}, V^{\prime}=\left(q_1, q_2, \ldots, q_H\right), \\
\left(k_1, k_2, \ldots, k_H\right),\left(v_1, v_2, \ldots, v_H\right) 
\end{multline}
\begin{multline}
\operatorname{MSSA}\left(Q^{\prime}, K^{\prime}, V^{\prime}\right)={\operatorname{ConvBN}} \\
\left(\operatorname{SN}\left(\left(q_1 k_1^T v_1 * s\right), \ldots,\left(q_H k_H{ }^T v_H{ }* s\right)\right)\right)
\end{multline}
where $Q, K, V \in \mathbb{R}^{T \times N\times D}$, and reshaped into H-head form
$Q^{\prime}, K^{\prime}, V^{\prime} \in \mathbb{R}^{T \times H \times N \times d}$ with $D = H \times d$. Note that the scaling factor $s$ in SSA or MSSA is a constant, and can be easily fused into the following spike neuron LIF layer.

\subsection{Spikingformer and Spikingformer$^\dagger$}\label{subsec:application}
\begin{table}[htb]
  \centering
  \vspace{-5mm}
  \caption{Experimental results of our proposed ConvBN-MaxPool-LIF Downsampling in CIFAR10/100, comparing with potential or mainstream downsampling way in SNN. In detail, we keep the retaining network structure of Spikingformer unchanged. Note that ConvBN-LIF-MaxPool, which is used in Spikformer, is the baseline.}
  \vspace{+2mm}
    \begin{tabular}{lccc}
    \toprule
    Method  & T & CIFAR10 & CIFAR100 \\
    \midrule
    ConvBN-LIF-MaxPool  & 4     & 95.81 & 79.21 \\
    \textbf{ConvBN-MaxPool-LIF}  & 4     & \textbf{95.95} & \textbf{80.37} \\
    ConvBN-AvgPool-LIF  & 4     &    95.23   & 78.52 \\
    ConvBN(stride=2)-LIF  & 4     &   94.94    & 78.65 \\
    \bottomrule
    \end{tabular}%
  \label{tab:downsamping}%
\end{table}%
Spikingformer$^\dagger$ is the Spikingformer with ConvBN-MaxPool-LIF (CML) downsampling. CML can overcome the imprecision problem of gradient backpropagation in Spikformer or Spikingformer, which improves the network performance while reduces the computational cost at the same time. The ablation study on CIFAR are shown in Tab. \ref{tab:downsamping}.

\subsection{Supplement of experimental details}
In our experiments, we use 8 GPUs when training models on ImageNet, while 1 GPU is used to train other datasets (CIFAR10, CIFAR100, DVS128-Gesture, CIFAR10-DVS). In addition, we 
adjust the value of membrane time constant $\tau$ in spike neuron when training models on DVS datasets. In direct training SNN models with surrogate function,
\begin{equation}
{Sigmoid}(x)=\frac{1}{1+\exp (-\alpha x)}
\end{equation}
We select the Sigmoid function as the surrogate function with $\alpha=4$ in all experiments.

\begin{figure*}[tbp]
	\centering
	\includegraphics[width=.99\textwidth]{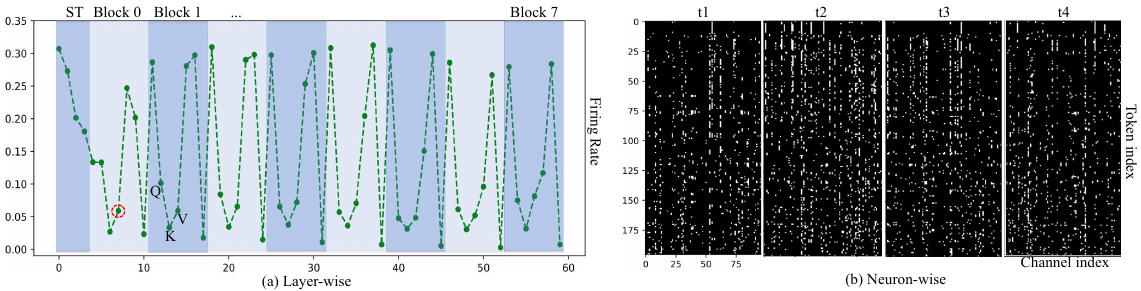}
	\caption{Firing Patterns. (a) Layer-wise firing rates of Spikingformer-8-768 on ImageNet. "ST" denotes the Spiking tokenizer. (b) Firing patterns of 196 $\times$ 96 neurons from layer 7 (red circle on the left) of Spikingformer-8-768 on ImageNet,  where the white dots represent firing. }
	\label{fig: Firing Patterns}
\end{figure*}

\begin{figure*}[tbp]
	\centering
	\includegraphics[width=.99\textwidth]{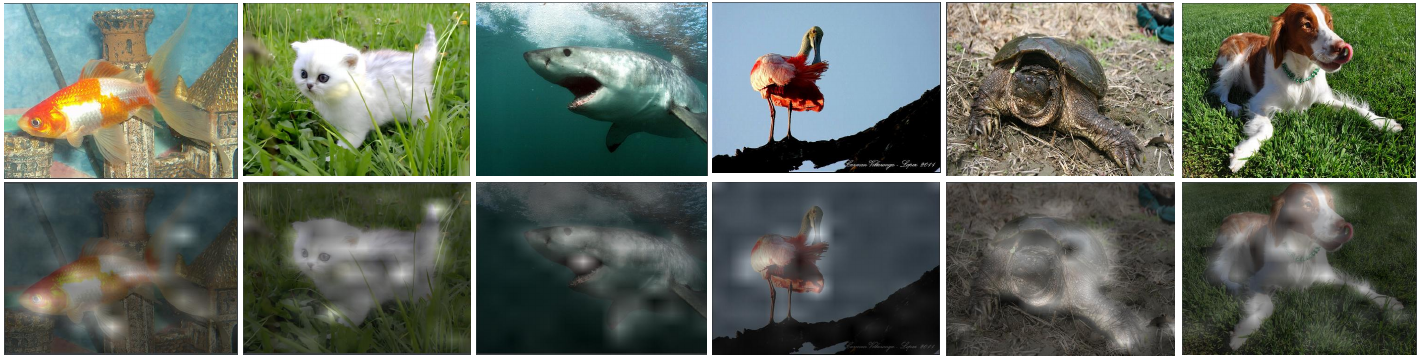}
	\caption{The attention map visualization of PSSA in Spikingformer.}
	\label{fig: Attention visualization}
	\vspace{-0.2cm}
\end{figure*}

\subsection{Energy Consumption Recalculation for Spikformer on ImageNet.}\label{sec:recalculation}
The energy consumption of Spikformer treat the non-spike computation (integer-float multiplications) in $\operatorname{ConvBN}$ layer as binary spike-based accumulate operations. That is obviously unreasonable. Therefore, we provide two ways to recalculate the energy consumption of Spikformer on ImageNet: 1) treating integers $N$ ($N>1$) multiplicating with floats as N times binary spike-based accumulate operations; 2) treating integers $N$ multiplicating with floats as floating point multiplication, which leads to much higher energy consumption. The recalculation in this work is according to the first way. 

\subsection{Further Analysis for Last Layer}\label{sec:Last Layer}
We carry out an analysis of the last layer of Spikingformer in CIFAR10 and CIFAR100 dataset to study its impacts, and the results are shown in Tab.\ref{tab:last layer}.
The last layer has a significant impact on model performance, although it only occupies a very small component in Spikingformer. 
Our experiments show that Spikingformer-$L$-$D^*$ with spike signals performs worse than Spikingformer-$L$-$D$ on the whole. This can be attributed to the fact that the $\operatorname{AvgPooling}$ layer of $\operatorname{AvgPooling}\text{-}\operatorname{FC}$ processes floating point numbers, which have stronger classification abilities than spike signals in $\operatorname{SN}\text{-}\operatorname{AvgPooling}\text{-}\operatorname{FC}$. 
However, Spikingformer-$L$-$D^*$ still outperforms Spikformer-$L$-$D$, which uses a $\operatorname{SN}\text{-}\operatorname{AvgPooling}\text{-}\operatorname{FC}$ layer as the classifier.  On one hand, Spikingformer effectively avoids integer-float multiplications of Spikformer in residual learning. On the other hand, Spikingformer has better spike feature extraction and classification ability than Spikformer. These results further verify the effectiveness of Spikingformer as a backbone.
\begin{table}[htb]
  \centering
  \caption{Discussion results on the last layer of Spikingformer. Spikingformer-$L$-$D$$^*$ means Spikingformer with the last layer of $\operatorname{SN}\text{-}\operatorname{AvgPooling}\text{-}\operatorname{FC}$. Spikingformer-$L$-$D$ means Spikingformer with the last layer of $\operatorname{AvgPooling}\text{-}\operatorname{FC}$ by default.}
  \vspace{+2mm}
    \begin{tabular}{llp{0.4cm}<{\centering}p{0.85cm}<{\centering}}
    \toprule
    Dataset & \multicolumn{1}{c}{Models} & Time Step & Top-1 Acc \\
    \midrule
    \multirow{3}{*}{CIFAR10} 
          & Spikingformer-4-384-400E  & 4     & 95.81  \\
          & Spikingformer-4-384-400E$^*$   & 4     & 95.58  \\
          & Spikformer-4-384-400E & 4     & 95.51 \\
    \midrule
    \multirow{3}{*}{CIFAR100} 
          & Spikingformer-4-384-400E  & 4     & 79.21 \\
          & Spikingformer-4-384-400E$^*$   & 4     &  78.39 \\
          & Spikformer-4-384-400E & 4     &  78.21 \\
    \bottomrule
    \end{tabular}%
  \label{tab:last layer}%
\end{table}%

\subsection{Additional Results on CIFAR10}
We trained Spikingformer$^\dagger$ on CIFAR10 up to 600 epochs, and the accuracy could increase up to 96.14$\%$. 
\begin{table}[htbp]
  \centering
  \vspace{-5mm}
  \caption{Training Spikingformer$^\dagger$-4-384 up to 600 epochs.}
  \vspace{+2mm}
    \begin{tabular}{cccc}
    \toprule
    Backbone & Epochs & Timestep & CIFAR10 \\
    \midrule
    \multicolumn{1}{l}{\multirow{3}{*}{\textbf{Spikingformer$^\dagger$}}}
    &300E & 4     & 95.81 \\
    &400E & 4     & 95.95 \\
    &600E & 4     & \textbf{96.14} \\
    \bottomrule
    \end{tabular}%
  \label{tab:cifar600e}%
\end{table}%

\subsection{Additional analysis on GLUE.}
We pretrain Spikingformer only on Wikipedia-English \citep{devlin2019bert} using masked language modeling \citep{devlin2019bert} with 8 GPUs, and subsequently fine-tune it on the GLUE dev set.
In this task, we retained the softmax layer in the attention map computation, consistent with SpikeLM \cite{xing2024spikelm}. We find this layer is crucial for model convergence in language tasks, which is different from visual tasks. 
In comparison, SpikeLM \cite{xing2024spikelm} is pretrained on both BooksCorpus \citep{zhu2015aligning} and Wikipedia-English \citep{devlin2019bert}, and retains the non-spiking GeLU activation in all MLP blocks, both of which contribute to its superior performance.

\subsection{Dataset and Training Details.}
\textbf{ImageNet} contains around $1.3$ million $1000$-class images for training and $50,000$ images for validation. The input size of our model on ImageNet is set to the default $224\times 224$. 
The optimizer is AdamW and the batch size is set to $192$ or $288$ during $310$ training epochs with a cosine-decay learning rate whose initial value is $0.0005$. The scaling factor is $0.125$ when training on ImageNet and CIFAR.
Four SPEDs in Spiking Tokenizer splits the image into $196$ $16 \times 16$ patches. 

\textbf{CIFAR10 / CIFAR100} provides 50, 000 train and 10, 000 test images with 32 × 32 resolution. The difference is that CIFAR10 contains 10 categories for classification, but CIFAR100 contains 100 categories, owning better distinguishing ability for classification algorithm. The batch size of Spikingformer is set to 64. We choose two SPEs and two SPEDs in the Spiking Tokenizer block to split the input image into 64 4 × 4 patches.

\textbf{CIFAR10-DVS} is a neuromorphic dataset converted from the static image dataset by shifting image samples to be captured by the DVS camera, which provides 9, 000 training samples and 1, 000 test samples.

\textbf{DVS128-Gesture} is a gesture recognition dataset that contains 11 hand gesture categories from 29 individuals under 3 illumination conditions. The image size of DVS128-Gesture is 128*128. The main hyperparameter setting in DVS128-Gesture classification is the same with CIFAR10-DVS classification. The only difference is that the number of training epoch is set as 200 for DVS Gesture classification, which is the same with Spikformer.

\textbf{General Language Understanding Evaluation (GLUE)} benchmark is a widely used evaluation framework designed to assess the performance of natural language understanding (NLU) models across a diverse set of language tasks. It cover a range of tasks, including single-sentence classification, sentence pair classification, and linguistic acceptability. Due to its diversity and task difficulty, GLUE has become a standard benchmark for evaluating language models. Specifically, our experiments focus on the following eight datasets within GLUE:

MNLI (Multi-Genre Natural Language Inference): A natural language inference task where models predict whether a given hypothesis is entailed by, contradicts, or is neutral to a premise, using examples from multiple genres.

QQP (Quora Question Pairs): A paraphrase detection task that requires identifying whether two questions from Quora have the same meaning.

QNLI (Question Natural Language Inference): A question-answering classification task reformulated as natural language inference, where models determine if a sentence contains the answer to a question.

SST-2 (Stanford Sentiment Treebank): A binary sentiment classification task focused on predicting the sentiment (positive/negative) of movie reviews.

CoLA (Corpus of Linguistic Acceptability): A binary classification task that determines whether a given sentence is grammatically acceptable according to linguistic judgments.

STS-B (Semantic Textual Similarity Benchmark): A regression task that measures the semantic similarity between two sentences on a scale from 0 to 5.

MRPC (Microsoft Research Paraphrase Corpus): A binary classification task for detecting whether two sentences are paraphrases of each other.

RTE (Recognizing Textual Entailment): A binary classification task where models predict whether a hypothesis can be inferred from a given premise, using datasets from various entailment challenges.

These eight tasks collectively evaluate key aspects of language understanding, including paraphrase detection, sentiment analysis, grammaticality, and semantic similarity, providing a comprehensive benchmark for language models.

\subsection{Attention Visualization on Spikingformer}
In this part, we visualized the firing patterns and Pre-activation Spiking Self-Attention in Spikingformer. The results are shown in Fig.\ref{fig: Firing Patterns} and Fig.\ref{fig: Attention visualization}, respectively.

\end{document}